\pdfoutput=1

\documentclass[11pt]{article}
\usepackage{acl2023}
\usepackage{float}
\usepackage{threeparttable}
\usepackage{lipsum}
\usepackage{stfloats}
\usepackage{amssymb}
\usepackage[inline]{enumitem} 
\usepackage{bm}
\usepackage{mathtools}
\usepackage{multirow}
\usepackage{makecell}
\usepackage{bbm}
\usepackage{multirow}
\usepackage{xcolor}

\usepackage{xparse}

\usepackage{booktabs}
\makeatletter
\newcommand\footnoteref[1]{\protected@xdef\@thefnmark{\ref{#1}}\@footnotemark}
\makeatother

\usepackage{times}
\usepackage{latexsym}

\usepackage[T1]{fontenc}

\usepackage[utf8]{inputenc}
\usepackage{amsmath}
\usepackage{tabularx}
\usepackage{microtype}

\title{Human in the loop: How to effectively create coherent topics by manually labeling only a few documents per class}

\author{Anton Thielmann \and Benjamin Säfken  \\
        Chair of Data Science and Applied Statistics \\ TU Clausthal\\ 
        \And Christoph Weisser \\
  Chair of Statistics \\ University of Göttingen \\}

\begin{document}
\maketitle

\begin{abstract}
Few-shot methods for accurate modeling under sparse label-settings have improved significantly. However, the applications of few-shot modeling in natural language processing remain solely in the field of document classification. With recent performance improvements, supervised few-shot methods, combined with a simple topic extraction method pose a significant challenge to unsupervised topic modeling methods.  Our research shows that supervised few-shot learning, combined with a simple topic extraction method, can outperform unsupervised topic modeling techniques in terms of generating coherent topics, even when only a few labeled documents per class are used. 
\end{abstract}

\setcounter{page}{1}

\section{Introduction}
The identification of latent topics in large text corpora has undergone a great deal of development. However, uncovering the hidden semantics of large text corpora is still, if not of ever-increasing interest. Scientific methods continue to evolve and achieve increasingly impressive results in terms of topic coherence \citep{larochelle2012neural, srivastava2017autoencoding, chien2018latent, wang2019atm, dieng2020topic}.
The practical relevance of such methods is evident from the large number of practical application papers alone. Topic models for information extraction are used, for example, for applied research in education \citep{granic2019technology}, offsite construction \citep{liu2019trending}, bioinformatics \citep{liu2016overview}, communication sciences \citep{maier2018applying} and many other practical areas (e.g. \citep{hall2008studying, daud2010knowledge, boyd2017applications, jelodar2019latent, hannigan2019topic}).

While all of these methods take an unsupervised approach, few-shot methods achieve remarkable results in various supervised label-scarse settings. The metrics of interest are in this case not the coherence of clusters, but model accuracy, F1 score, or precision. Huggingfaces Sentence Transformer Finetuning (\textsc{SetFit}) \citep{tunstall2022efficient} allows for such a small amount of labeled data while achieving impressive classification results that unsupervised methods are heavily challenged.
 
 The idea is simple. When less and less labeled documents per class are necessary for supervised methods to achieve state-of-the-art results, human input by manually labeling a few documents becomes an attractive option for unsupervised tasks such as document clustering. By leveraging pre-trained sentence transformers \citep{reimers-2019-sentence-bert} and class-based term frequency inverse document frequency (tf-idf) for topic extraction, we can generate coherent topics with only a few labeled documents per class. As a result, manually labeling a training data set and subsequently leveraging \textsc{SetFit} reduces the tiresome and time- and money-intensive manual labeling by such a dramatic amount, that it is a viable alternative to unsupervised approaches.

\paragraph{Contributions} The contributions of the paper can be summarized as follows:

\begin{enumerate}
    \item We present a method for \textbf{D}ocument \textbf{C}lassification and subsequent \textbf{T}opic \textbf{E}xtraction (DCTE) based on \textsc{SetFit}. The proposed method generates coherent topics from only a few labeled documents.
    \item We conduct a benchmark study, comparing the proposed approach with state-of-the-art topic models and document clustering methods.
    \item We outperform competitive benchmark models on three standard datasets in terms of topic coherence and create informative topics.
\end{enumerate}

\section{Related Work and Background}

Generative probabilistic models inspired by Latent Dirichlet Allocation (LDA) \cite{blei2003latent} topic models are still widely used. Several extensions, heavily drawing from LDA and also leveraging word-embeddings achieve state of the art results in terms of coherence \cite{wang2019atm, dieng2020topic}. Based upon \citet{srivastava2017autoencoding}, \citet{bianchi-etal-2021-cross} introduce the Zero-shot Topic model, which enables zero-shot cross lingual tasks.
Word-, document- and sentence- embeddings generated such a great performance impact, that even structurally simple models, leveraging pre-trained word- and sentence-embeddings and clustering these embeddings achieve remarkable results \cite{grootendorst2022bertopic, sia2020tired, angelov2020top2vec}. The embedding types range from doc2vec \cite{le2014distributed} over sentence-transformers \citep{reimers-2019-sentence-bert, reimers-2020-multilingual-sentence-bert} to word-embeddings generated with BERT \cite{devlin2018bert, liu2019roberta}. Modeling techniques include K-Means, Hierarchical Density-Based Spatial Clustering of Applications with Noise (HDBSCAN) \cite{mcinnes2017hdbscan} and Gaussian Mixture Models (GMM) \cite{reynolds2009gaussian}. Topics are extracted with class based tf-idf \cite{grootendorst2022bertopic} or based on distance measures in the feature space \cite{angelov2020top2vec, sia2020tired}. Leveraging pre-trained embeddings from large scale language models seems to positively impact modeling performance \cite{bianchi2020pre}.

The inclusion of labeled data into topic modeling was mostly done to improve unsupervised results in text mining \citep{ramage2011partially} and for multi-labeled corpora \citep{ramage2009labeled}. Few-shot topic modeling has only been of interest as of late \citep{iwata2021few, duan2022bayesian}. \citet{iwata2021few} introduces a few-shot model that relies on a neural network generating priors for generative probabilistic topic modeling and achieves impressive results with respect to perplexity. \citet{duan2022bayesian} introduces a bi-level generative model combined with a topic-meta learner. However, both few-shot methods are designed to create great topics from little data. A setting that is exceedingly unlikely given the ever growing availability of large text corpora. In contrast, leveraging existing few-shot methods for generating coherent topics for very small samples of labeled training data has a high practical relevance. When only a handful of documents have to be labeled manually to improve topic coherence, creating humanly labeled training data is a viable and effective option.

\paragraph{\textsc{SetFit}} The leveraged model, \textsc{SetFit} \citep{tunstall2022efficient}, can be described as a two-step algorithm. In a first step, an already pre-trained Sentence Transformer \cite{reimers-2019-sentence-bert} is fine-tuned using only a few-labeled samples per class. A siamese, contrastive architecture is used on sentence/document pairs to ensure better generalizability. Creating these contrastive learning pairs artificially enlarges the training data set. The size of this fine-tuning dataset is therefore dependent on the number of labeled training sentences and classes. For $k$ classes and $n$ equally distributed samples per class, we can hence construct $\sum_{i=1}^{k-1} n (k-i) n$ contrastive learning pairs. Each contrastive learning pair thus consist of one positive sample from class $c$ and one randomly selected negative sample from a different class. This contrastive architecture increases the small amount of training data by a margin and enables the model to achieve the impressive classification results with extremely little data as shown by \citet{tunstall2022efficient}.
Second, a classification head is trained using the encoded training corpus. 

\section{Methodology}
Our proposed methodology is surprisingly simple, yet highly effective. 
Let the vocabulary of words be expressed as $\textup{V} =\left \{ w_1,~\ldots~, w_n   \right \}$. Let the corpus, i.e. the collection of documents be expressed as $\textup{D} =\left \{ d_1,~\ldots~, d_M   \right \}$.
Further, let each document be expressed as a sequence of words $d_i = \left [ w_{i1}, \ldots, w_{in_i} \right ]$  where $w_{ij} \in V$ and $n_i$ denotes the length of document $d_i$.  $\mathcal{D} = \left \{ \bm{\delta}_1,~\ldots~, \bm{\delta}_M   \right \}$ then denotes the set of documents represented in the embedding space, such that $\bm{\delta}_i$ is the vector representation of $d_i$. Further, let a topic $t_k$ from a set of topics $ T = \left \{ t_1,~\ldots~, t_K   \right \}$ be represented as a discrete probability distribution over the vocabulary \citep{blei2003latent}, such that $t_k$ is expressed as $(\phi_{k, 1}, \ldots, \phi_{k, n} )^T$ and $\sum_{i=1}^{n} \phi_{k, i}=1$ for every $k$.

\begin{figure}
    \centering
        \includegraphics[scale=0.3]{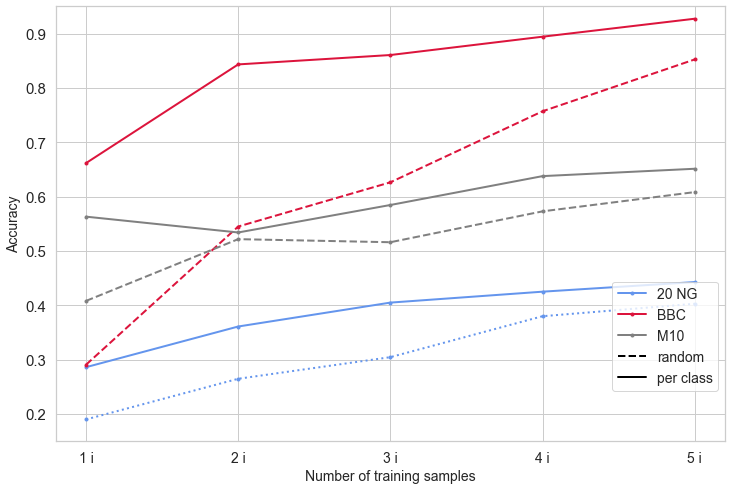}
        \caption{\small{The models average classification accuracies dependent on the number of labeled training samples. As expected, the accuracy increases with the number of labeled samples. i denotes the number of classes present in the dataset. Each model  is fit 5 times, using different randomly selected documents in the training corpus. We find that a model's coherence and a model's accuracy are independent from another (see Table \ref{samples_coh}).}}
    \label{fig:my_label}
\end{figure}

To generate topics, we fine-tune a Sentence-Transformer \cite{reimers-2019-sentence-bert} leveraging the \textsc{SetFit} architecture. Subsequently, we train a classification head on the fine-tuned document embeddings $\mathcal{D}$. Using a Neural Network and bypassing classical unsupervised clustering allows to circumvent dimensionality reduction as opposed to \citet{sia2020tired, grootendorst2022bertopic} or \citet{angelov2020top2vec}. As Deep neural networks are not susceptible to a dimensionality curse, no information is lost during a dimensionality reduction step. The classification head is trained only with the few labeled samples. The remaining documents are mapped into the embedding space and classified. Subsequently, the topics are extracted from the resulting document clusters.

\paragraph{Topic Extraction}
As the proposed method only results in document clusters, but not a set of topics $T$, we must extract the topics from the document clusters. 
We use a method already proven successful in the literature, the class-based tf-idf approach \citep{salton1989automatic, grootendorst2022bertopic},
\begin{equation*}
    \mbox{tf-idf} (w | c) = \frac{frequency(w_c)}{n_c} \cdot \log \left(\frac{N}{\sum_j w_j}\right).
\end{equation*}
\noindent With $frequency(w_c)$ being the total frequency of the word in class $c$, $n_c$ being the total number of words in class $c$, N being the total number of documents and $\sum_j w_j$ being the overall frequency of word $w$ over all classes. A topic, $t_k$ is hence represented by the top $j$ words according to the words normalized \textit{tf-idf} scores.

\begin{table*}
\centering
\fontsize{8}{9.5}\selectfont{
    \begin{tabular}{l | l| c  c  c  || l |  c  c  c }
     & \multicolumn{4}{c}{\textbf{Random draw}}  & \multicolumn{4}{c}{\textbf{Per class}}\\
     & Samples & Average & Max & Std. &  Samples & Average & Max & Std. \\
      \hline
     \multirow{5}{1cm}{20 News- \\ groups} & 20    & \bf{0.185} &  \bf{0.221}  &  $\pm$0.080 & 1 &  0.117 & 0.163 & $\pm$0.036  \\
                              & 40    & 0.108 &  0.144  &  $\pm$0.024 & 2 &  0.115 & 0.196 & $\pm$0.046  \\
                              & 60    & 0.122 &  0.190  &  $\pm$0.032 & 3 &  0.137 & 0.189 & $\pm$0.061  \\
                              & 80    & 0.145 &  0.190  &  $\pm$0.045 & 4 &  \bf{0.186} & \bf{0.208} & $\pm$0.021  \\
                              & 100   & 0.162 &  0.200  &  $\pm$0.047 & 5 &  0.164 & 0.192 & $\pm$0.027  \\
      \bottomrule
    \multirow{5}{1cm}{\textit{BBC News}}    & 5     & 0.097 &  \bf{0.20}  &  $\pm$0.084  & 1 &  0.103 & 0.153 & $\pm$0.032  \\
                              & 10    & \bf{0.192} &  0.152  &  $\pm$0.059  & 2 &  0.133 & 0.187 & $\pm$0.047  \\
                              & 15    & 0.115 &  0.124  &  $\pm$0.054  & 3 &  0.107 & 0.180 & $\pm$0.041  \\
                              & 20    & 0.081 &  0.139  &  $\pm$0.024  & 4 &  0.117 & \bf{0.191} & $\pm$0.040  \\
                              & 25    & 0.121 &  0.115  &  $\pm$0.020  & 5 &  \bf{0.142} & 0.186 & $\pm$0.022  \\
      \bottomrule
    \multirow{5}{1cm}{M10}  & 10    & -0.178 & \bf{-0.015}   & $\pm$0.119  & 1 & -0.146  & -0.078 & $\pm$0.037  \\
                             & 20    & -0.153 & -0.12    & $\pm$0.021  & 2 & \bf{-0.115}  & \bf{-0.054} & $\pm$0.039  \\
                             & 30    & -0.142 & -0.078   & $\pm$0.033  & 3 & -0.158  & -0.107 & $\pm$0.032  \\
                             & 40    & -0.098 & -0.033   & $\pm$0.055  & 4 & -0.119  & -0.103 & $\pm$0.015  \\
                             & 50    & \bf{-0.094} & -0.054   & $\pm$0.035  & 5 & -0.121  & -0.098 & $\pm$0.027  \\
      \end{tabular}
      }
        \caption{\small{Experimental results for different numbers of labeled training samples. The average NPMI coherence and standard deviations over  5 runs is presented. All models are fit using the \textit{all-MiniLM-L6-v2} sentence transformer. The biggest coherence score for each column for each dataset is marked in bold.}}
    \label{samples_coh}
\end{table*}

\section{Experiments}

\paragraph{Evaluation}
To evaluate the topics, we use normalised pointwise mutual information (NPMI) coherence scores \citep{lau2014machine}. 
\begin{equation*}
    NPMI(t_k) = \sum_{j=2}^{N}\sum_{i=1}^{j-1} \frac{log\frac{(P(w_i, w_j)}{P(w_i) P(w_j)}}{-log(P(w_i, w_j)}
\end{equation*}
We use the training corpora as the reference corpora for constructing coherence scores respectively. Stopwords are removed to punish models that include meaningless words into their topics more severely and not favor models constructing information-less topics with frequently co-occurring words. $N$ is set to 10 for all evaluations over all models.

\paragraph{Experimental setup}
We use the \textit{20 Newsgroups}, \textit{BBC News} and \textit{M10} corpora as the benchmark datasets.
To circumvent any induced bias in the results by a lucky selection of labeled training documents, we train the model several times, each time with different randomly selected labeled training documents. 
All topics are evaluated with coherence scores. We compare the results with state-of-the-art unsupervised topic modeling and document clustering approaches. To achieve the best possible comparability we choose the same pre-trained sentence-transformer, \textit{all-MiniLM-L6-v2} \citep{reimers-2019-sentence-bert}, for all benchmark models where applicable\footnote{As comparison models, we use BERTopic \citep{grootendorst2022bertopic} as a representative of clustering based topic models, LDA \citep{blei2003latent} as a model not leveraging pre-trained embeddings, CTM \citep{bianchi-etal-2021-cross} as a generative probabilistic model leveraging pre-trained embeddings, a simple K-Means model - closely following the architecture from \citet{grootendorst2022bertopic}, but replacing HDBSCAN with a K-Means clustering approach, ETM \citep{dieng2020topic} leveraging word2vec \citep{mikolov2013efficient} and NeuralLDA and ProdLDA \citep{srivastava2017autoencoding}}. All models are fit using the OCTIS framework \citep{terragni2021octis}. A detailed description of the models hyperparameters and the hyperparameter tuning can be found in the Appendix, \ref{hyper_tuning}.

For DCTE we compare two different labeling frameworks. One, which uses $n=1, \dots, 5$ labeled randomly drawn documents per class and one where we use $i$ randomly labeled documents without correcting for true labels, with $i$ being dependent on the dataset. $i = 20, 40, \dots, 100$ for \textit{20 Newsgroups}, $i = 5, 10, \dots, 25$ for \textit{BBC News} and $i = 10, 20, \dots, 50$ for \textit{M10}.
Note, that we perform hyperparameter tuning for all benchmark models (see Appendix, \ref{hyper_tuning}), but use a vanilla approach for DCTE. This demonstrates a useful real-world applicability as the method generates coherent topics \textit{out of the box} and independent of the dataset or even the selected training samples (see Table \ref{samples_coh}).

\paragraph{Results}

The results for the proposed approach can be seen in table \ref{samples_coh}. 
We find that even with a small amount of labeled data, coherent topics can be constructed. The coherence does not depend on the number of training samples and one labeled document per class is sufficient to create coherent topics. When randomly drawing from the corpus, we find larger standard deviations in the average coherence scores over 5 runs. This is due to the fact that with random sampling, one might fail to include all classes present in the dataset into the training data. This is also represented by the poor classification accuracies represented in Figure \ref{fig:my_label}. However, we find that even under these conditions, coherent topics are created.
Table \ref{benchmark} shows the performance compared to the benchmark models. Additionally, we find that the presented method not only creates coherent but also informative topics (See Appendix, tables \ref{tab:example comparison} - \ref{tab:example topics_ctm}). 
Even with these few labeled training samples per class, DCTE creates coherent topics. The use of supervised methods and especially the lack of any form of dimensionality reduction seem to have a positive effect on topic coherence.

\section{Conclusion}
In this paper, we show that recent improvements in few-shot models make manual labeling of a few training documents a valid alternative to unsupervised topic modeling. With a small amount of human input in the form of labeled training samples, even simple topic extraction methods can yield great results in terms of topic coherence.
Additionally, the achieved coherence scores are achieved without any form of hyperparameter tuning and relatively low number of epochs and text pairs for the contrastive learning (Appendix, \ref{hyper_tuning}). While already achieving great coherence scores, this leaves further possibilities for improvement in the presented method. Furthermore, we find that the created document-topic distributions  match the underlying distributions in the dataset (Appendix, \ref{comp}). 

\begin{table}[H]
\small
\begin{threeparttable}

    \begin{tabular}{p{1.8cm}|p{1.3cm} | p{1.4cm} | p{1.3cm}}
     & \multicolumn{3}{c}{NPMI}  \\ 
    Model & {\textit{20 News}} & {\textit{BBC News}} & {\textit{M10}} \\
        \hline\hline
        LDA & 0.096 & -0.214 & -0.218 \\
        NeuralLDA & 0.046 & -0.357 & -0.55\\
        ProdLDA & 0.161 & -0.099 & \textbf{-0.09}  \\
        BERTopic & -0.10 & 0.044 & -0.303 \\
        BERTopic\tnote{*} & 0.128 & \textbf{0.2068} & -0.126  \\
        K-means & 0.115 & 0.0648 & -0.134 \\
        ETM & -0.089 & -0.077 & -0.188 \\
        CTM & \textbf{0.205} & -0.002 & -0.213 \\
        \midrule
        DCTE\tnote{1}     &      \textbf{0.221} & \textbf{0.20} &  \textbf{-0.015} \\  
        DCTE\tnote{2}     &      \textbf{0.163} & \textbf{0.153} & \textbf{-0.054}  \\
        DCTE\tnote{3}     &    0.117 & 0.103 & -0.146 \\
        DCTE\tnote{4}     &    \textbf{0.186} & \textbf{0.117} & \textbf{-0.119} \\
        \bottomrule
    \end{tabular}
     \begin{tablenotes}
      \small
      \begin{enumerate*}
      
      \item[*] Only Evaluating the top 50\% coherent topics. \\ 
      \item[$^{1}$] The most coherent model, using 20, 5 and 10 randomly drawn labeled training samples respectively. \\ 
      \item[$^{2}$] The best model achieved with only one labeled training sample per class. \\ 
      \item[$^{3}$] The average achieved coherence when using only one labeled training sample per class.\\
      \item[$^{4}$] The average achieved coherence when using 4 labeled training samples per class.
      \end{enumerate*}
      
    \end{tablenotes}
    \end{threeparttable}
        \caption{\small{NPMI coherence scores for all tested models on the three benchmark datasets. CTM, BERTopic, K-Means and \textsc{SetFit} are all fit using \textit{all-MiniLM-L6-v2}  \citep{reimers-2019-sentence-bert}. LDA is fit using standard bag-of-words representations. NeuralLDA and ETM are fit using Word2Vec \citep{mikolov2013efficient}} As BERTopic uses HDBSCAN, it detects the number of topics automatically. However, it drastically overestimates the number of true topics for all datasets. To account for garbage topics negatively impacting the coherence scores, we favorably choose to also evaluate only the top 50 \% coherent topics from that output.
        The top four coherent models are marked in bold.}
       
    \label{benchmark}
\end{table}

\newpage
\section{Limitations}
While there are multiple advantages of the presented method there are also apparent limitations. Although we can effectively generate coherent topics with a very small amount of labeled data, it still requires manual labeling. Additionally, manual labeling requires an idea of how many different topics are present in the corpus. However, most unsupervised methods also require setting a fixed number of topics \cite{blei2003latent, sia2020tired, dieng2020topic}. Moreover, most algorithms optimize the number of extracted topics over coherence scores, which could easily be done with the presented method. Note that the presented benchmarks for DCTE are all achieved without any form of hyperparameter tuning. This reduces computation time and artificially creates an idea of how real-world applications could benefit from this method. This paper does not delve into results from different few-shot document classifiers, (e.g.\citep{rios2018few, pan2019few, cao2020concept}) or different pre-trained sentence transformers for document embeddings \citep{reimers-2019-sentence-bert}. Future research may implement additional few-shot methods to potentially find even better suited few-shot classifiers and embedding models for creating coherent topics.

Besides, additional applications could include multi-labeled and multi-topic documents.  Both of these problems can be easily solved with the presented method. The topic extraction method could be replaced by a similar method as used by \citet{sia2020tired}. Topical centroids could be constructed and distance measures in the embedding space could be used to extract the topics (see Appendix, \ref{extraction2}).

\newpage 
\bibliography{bib.bib}
\bibliographystyle{acl_natbib}

\appendix
\section{Appendix}
\paragraph{Topic Extraction} \label{extraction2}
Another method for topic extraction could be adapted from \citet{angelov2020top2vec} and \citet{sia2020tired}. In this approach, document clusters are first represented as topical centroids in the embedding space, $\bm{\mu}$. This allows for soft-clustering and hence multi-labeled documents, but requires additional computation steps and can be susceptible to a chosen embedding model. Second, the vocabulary $\textup{V} =\left \{ w_1,~\ldots~, w_n   \right \}$ is also mapped into the same feature space, such that $\mathcal{W} = \left \{ \bm{\omega}_1,~\ldots~, \bm{\omega}_n   \right \}$.
Hence, each word $w_i$ in the embedding space represented as $\bm{\omega}_i  \in \mathbb{R}^{L}$ has the same dimensionality $L$ as a document vector $\bm{\delta}_i  \in \mathbb{R}^{L}$.
There are two ways to represent a document as an average over word-embeddings. First, using the approach from \citet{sia2020tired}, which involves representing a document as an average of word-embeddings created with BERT \citep{devlin2018bert}. Second, interpreting the vocabulary as one-word sentences and using the same embedding model as for the documents. Subsequently, the similarity between every word and every topical centroid is computed e.g. as:
\begin{equation*}
    sim(\bm{\omega}, \bm{\mu}) = \frac{\bm{\omega} \cdot \bm{\mu}}{\lVert \bm{\omega}\rVert \Vert\bm{\mu}\lVert}, 
\end{equation*}

\noindent where $$\bm{\omega} \cdot \bm{\mu}=\sum_{i=1}^{L} \omega_{i} \mu_{i}$$ and $$\lVert\bm{\omega}\rVert  \lVert\bm{\mu}\rVert=\sqrt{\sum_{i=1}^{L} (\omega_i)^2}  \sqrt{\sum_{i=1}^{L} (\mu_i)^2}.$$ $L$ denotes the vectors dimension in the feature space, which is identical for $\bm{\omega}$ and $\bm{\mu}$.

\section{Experimental Details}\label{hyper_tuning}
All benchmark models are fitted 5 times for each dataset. The reported coherence scores are favorably the maximum coherence score achieved of the model during the 5 runs.

All benchmark models are fitted using the same pre-trained Sentence Transformer, all-MiniLM-L6-v2 \citep{reimers-2019-sentence-bert}, when applicable. The dimensionality reduction of the embedding in BERTopic \citep{grootendorst2022bertopic} is set as the default. Hence, the embeddings are reduced to 5 dimensions using umap \citep{mcinnes2018umap}. As intended by the author, HDBSCAN \citep{mcinnes2017hdbscan} is used for clustering. Because the number of clusters is detected automatically in HDBSCAN and we find that BERTopic heavily overestimates the number of true classes in the dataset, we report both, first the average coherence score over all topics and second the average coherence score for the top 50\% coherent topics. For the K-Means application, we closely follow the approach by \citet{grootendorst2022bertopic}, but change HDBSCAN to the K-Means algorithm, such that we can fix the number of topics manually. For dimensionality reduction, we optimize with respect to coherence scores and test a range from 2, to 20, using umap. We use 15 dimensions for all three datasets, as 15 dimensions performed marginally favorably compared to the 5 dimensions used in BERTopic \citet{grootendorst2022bertopic} and \citet{angelov2020top2vec}. We additionally tested the Top2Vec model \citep{angelov2020top2vec}, but as the results were inferior to BERTopic and they are very similar in the methodology we did not include it in the benchmark study.
For ETM \citep{dieng2020topic}, ProdLDA \citep{srivastava2017autoencoding} and NeuralLDA \citep{srivastava2017autoencoding} we train the word2vec embeddings simultaneously as intended by the authors. 

For CTM, ETM, ProdLDA and NeuralLDA, we iterate over a grid containing the following hyperparameters (when the hyperparameters are applicable to the respective model): Batch size, learning rate, dropout, hidden size, rho size, number of neurons, embedding size and the number of epochs.
Batch sizes are tested from 16 up to 1024 in factorials of 2. Learning rates from 2e-5 to 2e-1. Dropout is tested in steps of 0.1 from 0.1 to 0.8. Hidden sizes for ETM are tested from 500 to 1600 in steps of 200. Rho size for ETM is tested from 100 to 500 in steps of 100. The number of neurons are tested from 400 to 1600 in steps of 200, embedding size for ETM from 100 to 800 in steps of 100. The embedding size for ETM is tested from 100 to 500 in steps of 100. The number of epochs is tested from very few, 20 to 2000. Early stopping with the default Octis patience of 5 is implemented where applicable to the model.

For DCTE we use no hyperparameter tuning in order to simulate more relatedness to real-world applications. We train each model for 10 epochs with a learning rate of 2e-5. The number of contrastive learning pairs depends on the number of available samples to create contrastive learning pairs. Where possible, we use 10 contrastive learning pairs and otherwise the largest possible number.

\section{Training Data}
\paragraph{\textit{20 Newsgroups}}
For the \textit{20 Newsgroups} corpus, we reverse the classical train-test-split from scikit-learn. Hence, we randomly draw our training data from the scikit-learn (Pedregosa et al., 2011) test split. The scikit-learn training corpus is then predicted and the topics are extracted. All evaluation metrics are based upon 11,314 documents. All other models are hence fit on this training data. The dataset contains 20 topics\footnote{\textit{alt.atheism}, \textit{comp.graphics}, \textit{comp.os.ms-windows.misc}, \textit{comp.sys.ibm.pc.hardware}, \textit{comp.sys.mac.hardwa}, \textit{comp.windows.x}, \textit{misc.forsale}, \textit{rec.autos}, \textit{rec.motorcycles}, \textit{rec.sport.baseball}, \textit{rec.sport.hockey}, \textit{sci.crypt}, \textit{sci.electronics}, \textit{sci.med}, \textit{sci.space}, \textit{soc.religion.christian}, \textit{talk.politics.guns}, \textit{talk.politics.mideast}, \textit{talk.politics.misc}, \textit{talk.religion.misc}}. Some of these topics are very similar to one another, which explains DCTE's good coherence score, when only training with e.g. 15 classes.

A perfectly accurate model, precisely classifying each document correctly, would achieve a coherence score of 0.21 with the used class-based tf-idf topic extraction method. This is surprisingly lower, than the best DCTE model, which is only trained on 20 randomly drawn training samples.

Note, that the dataset is minimally preprocessed. We remove all stopwords, words that are shorter than 3 characters and strip punctuation and digits.

\paragraph{\textit{BBC News}}
For the \textit{BBC News} dataset, we again reverse the classical train-test split. However, we use the OCTIS \citep{terragni2021octis} implementation of the dataset. Hence, we sample the training documents for DCTE from the test dataset provided by OCTIS. The train and validation corpora are subsequently combined and used for the predictions. All other models are fitted on this training data. The complete dataset contains 2225 documents.
The dataset includes 5 topics: \textit{sport}, \textit{tech}, \textit{business}, \textit{entertainment} and \textit{politics}. The topics are relatively equally distributed: \textit{business}: 23\%, \textit{entertainment}: 17\%, \textit{politics}: 19\%, \textit{sport}: 23\%, \textit{tech}: 18\%.
OCTIS provides a preprocessed dataset, where stopwords, words containing less than 3 characters, punctuation and digits are removed.

A perfectly accurate model, precisely classifying each document correctly, would achieve a coherence score of 0.181 with the used class-based tf-idf topic extraction method. This is again a little bit lower than the best DCTE model. With only 2 labeled samples per class, coherence scores larger than 0.18 can be achieved with the presented method.

\paragraph{\textit{M10}}
For the \textit{M10} News dataset, we again reverse the classical train-test split. We use again the OCTIS \citep{terragni2021octis} implementation of the dataset. Hence, we sample the training documents for DCTE from the test dataset provided by OCTIS. The train and validation corpora are subsequently combined and used for the predictions. All other models are fitted on this training data. The complete dataset contains 8355 documents. The dataset includes 10 topics: \textit{agriculture}, \textit{archaeology}, \textit{biology},
\textit{computer science}, \textit{financial economics},\textit{ industrial engineering}, \textit{material science}, \textit{petroleum chemistry}, \textit{physics}, and \textit{social science}.
OCTIS again already provides the preprocessed dataset, including the same steps as described above

A perfectly accurate model, precisely classifying each document correctly, would achieve a coherence score of -0.102 with the used class-based tf-idf topic extraction method. This is considerably lower than for the first two datasets. This is also depicted by the coherence score DCTE and all benchmark models achieve which are all $<0$. However, the best DCTE model achieves considerably better coherence scores. Thus, having a perfectly accurate model might come at the cost of creating less coherent topics.

\section{Topic Analysis} \label{comp}
We heuristically show some topics created with DCTE and demonstrate reasonable document-topic distributions. Additionally, we compare some DCTE topics with topics created with the most coherent model from the benchmark.

\paragraph{\textit{20 Newsgroups}}
While models like CTM and ProdLDA achieve high topic coherences, we find that the models often create topics that contain little information. Table \ref{tab:example comparison} shows two exemplary topics. One created with the presented approach and one created with CTM. The CTM topic achieves a greater coherence score, but contains uninformative words like \textit{apparently} and \textit{frequently}.
\begin{table}[H]
    \centering
    \begin{tabular}{ l | c | c }
  Model  & CTM & DCTE \\ 
    \hline
 & success & patient\\ \ 
 & perform & health\\ 
 & initial & disease\\ 
 & complex & wire\\ 
 & aid & doctor\\ 
 & frequently & circuit\\
 & apparently & food\\ 
 & active & cancer\\ 
 & consist & ground\\ 
 & submit & use\\
\hline
NPMI & 0.216 & 0.102 \\ 
         
    \end{tabular}
    \caption{\small{Topic comparison between CTM and DCTE. The CTM topic achieves a higher coherence score, while being relatively uninformative. Multiple adverbs and non-case specific adjectives (e.g. \textit{complex}, \textit{active}) are included in the topic. The DCTE topic achieves a lower coherence score. Words like \textit{use} and \textit{ground} in the DCTE topic are also relatively uninformative. However, a large part of the topic represents a coherent \textit{medicine} topic.}}
    \label{tab:example comparison}
\end{table}

Table \ref{tab:example topics_ctm} contains two topics created with DCTE and a CTM. One labeled document per class was used during training for DCTE to create these topics. The topics \textit{Religion} and \textit{Space} are clearly distinguishable.

\begin{table}[H]
    \small
    \centering
    \begin{tabular}{ l | c | c || c | c }
    & \multicolumn{2}{c}{DCTE} & \multicolumn{2}{c}{CTM} \\
  Topic  & Religion & Space & Religion & Space \\ 
    \hline
& god & space           & conclusion  & year              \\ 
& jesus & launch        & science  & mission         \\ 
& church & satellite    & atheist  & launch         \\ 
& christian & mission   & church  & orbit           \\ 
& religion & orbit      & atheism  & solar        \\ 
& belief & moon         & truth  & make           \\ 
& word & data           & religion  & space            \\ 
& atheist & science     & tradition  & moon          \\ 
& faith & earth         & christian  & planet          \\ 
& people & rocket       & argument  & surface        \\
\hline
NPMI & 0.385 & 0.41 & 0.324 & 0.111  \\ 
         
    \end{tabular}
    \caption{\small{Topics created with DCTE and CTM  and the respective NPMI coherence scores. Both topics are clearly distinguishable and interpretable. Interestingly, CTM includes non-informative words as \textit{make} and \textit{year} in the \textit{Space} topic.}}
    \label{tab:example topics_ctm}
\end{table}

\begin{figure}[H]
    \centering
    \includegraphics[scale=0.28]{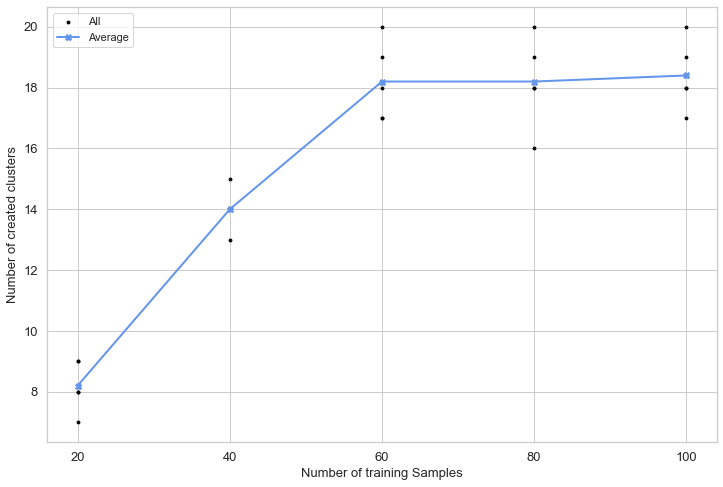}
    \caption{\small{Average number of extracted topics per labeled training samples for the \textit{20 Newsgroups} dataset. This is only for the completely randomly drawn labels. With a larger number of extracted training samples, the model is closer to the true number of classes present in the dataset.}}
    \label{fig:n_topics_20NG}
\end{figure}

When randomly drawing training samples from the corpus, we may fail to draw one document from each class. This is done to simulate real-world applications. While the accuracy of the model can subsequently suffer from that, the model still creates coherent topics. However, for the \textit{20 Newsgroups} dataset, the model created less than the actual 20 prevalent topics from the dataset. Figure \ref{fig:n_topics_20NG} show the dependency of the number of randomly drawn documents and created topics. As the dataset contains multiple topics that are very similar to one another, the number of extracted topics is still reasonable, which is also represented by the good coherence scores.

\begin{figure}[H]
    \centering
    \includegraphics[scale=0.5]{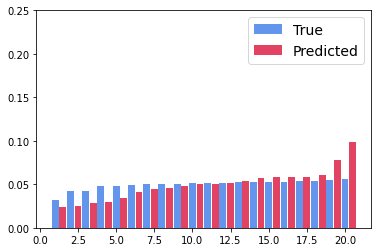}
    \caption{\small{Predicted topic distribution vs. true topic distribution. Results are achieved with DCTE on two labeled samples per class. Given the similarity of topics like \textit{mac.hardwarde} and \textit{pc.hardware} or \textit{religion.christian} and \textit{religion.misc} the achieved distribution, in combination with the achieved accuracies is reasonable.}}
    \label{fig:n_topicdist_20NG}
\end{figure}

To control for the model not creating arbitrarily large garbage topics and creating bad document-topic distributions, we check the created argmax predicion distribution against the true document topic distribution, see Figure \ref{fig:n_topicdist_20NG}. We find that although the model over and underestimates mainly 4 topics, the overall distribution does not suggest the creation of large garbage topics\footnote{\label{foot:note1} The distribution was created with two training samples per class and randomly selected from the 5 training runs.}.

\paragraph{\textit{BBC News}}
For the \textit{BBC News} dataset, most benchmark models struggle to identify the underlying latent topics. This is presumably due to the small amount of training data. As DCTE uses little training data anyway, the number of samples in a corpus does not effect the model's results. Table \ref{tab:example topics bbc} shows two exemplary topics from DCTE using one labeled training sample per class. The topics \textit{entertainment} and \textit{education} are clearly distinguishable.

\begin{table}[H]
    \centering
    \begin{tabular}{ l | c | c }
  Topic  & Entertainment & Education \\ 
    \hline
& film          & school           \\ 
& actor         & education         \\ 
& award         & child        \\ 
& actress       & teacher     \\ 
& star          & pupil      \\ 
& aviator       & student        \\ 
& director      & sport          \\ 
& role          & university       \\ 
& nomination    & parent         \\ 
& oscar         & democracy        \\ 
\hline
NPMI &  0.46 & 0.22 \\ 
         
    \end{tabular}
    \caption{\small{Topics created with DCTE and the respective NPMI coherence scores for the \textit{BBC News} dataset. The topics \textit{Entertainment} and \textit{Education} are clearly assignable.}}
    \label{tab:example topics bbc}
\end{table}

When randomly drawing training samples from the corpus, it might happen that we fail to draw one document from each class. This leads to the model detecting as many topics as we have classes in our training data. However, only labeling 25 documents already leads to 4.8 detected topics on average over 5 runs. As unsupervised methods require setting a fixed number of topics in advance, we believe that this is not a significant drawback of the model.
\begin{figure}[H]
    \centering
    \includegraphics[scale=0.3]{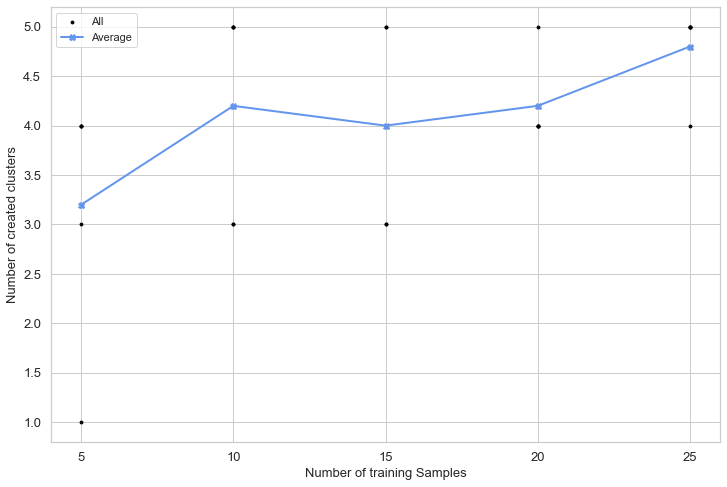}
    \caption{\small{Average number of extracted topics per labeled training samples. This is only for the completely randomly drawn labels.}}
    \label{fig:n_topics_bbc}
\end{figure}

We again analyze the document-topic distributions. For the \textit{BBC News} dataset all document classes are nearly equally present in the dataset, which is also captured by the presented method\footnoteref{foot:note1}.

\begin{figure}[H]
    \centering
    \includegraphics[scale=0.5]{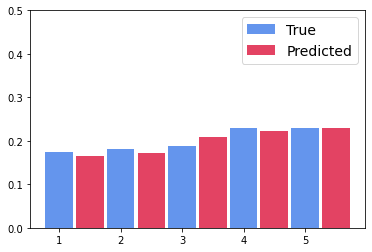}
    \caption{\small{Predicted topic distribution vs. true topic distribution. Results are achieved with DCTE on two labeled samples per class. Randomly drawing 10 samples from the dataset already leads to averagely more than 4 extracted topics.}}
    \label{fig:n_topicdist_BBC}
\end{figure}

\paragraph{\textit{M10}}
While the achieved coherence scores for the \textit{M10} dataset are not as good as for the other datasets, the created topics are still fairly well interpretable. Table \ref{tab:example topics M10} shows two topics created with the presented method and their respective coherence scores. The two topics, \textit{agriculture} and \textit{financial economics} present in the \textit{M10} corpus are clearly identifiable.

\begin{table}[H]
    \centering
    \begin{tabular}{ l | c | c }
  Topic  & Agriculture & Financial Economics \\ 
    \hline
& crop         & market       \\ 
& water        & stock        \\ 
& soil         & price   \\ 
& yield        & rate  \\ 
& climate      & volatility \\ 
& change       & option     \\ 
& irrigation   & return     \\ 
& management   & pricing       \\ 
& area         & risk     \\ 
& plant        & exchange       \\ 
\hline
NPMI &  0.155 &   0.191 \\ 
         
    \end{tabular}
    \caption{\small{Topics created with DCTE and the respective NPMI coherence scores for the \textit{M10} dataset. The topics \textit{Agriculture} and \textit{Financial Economics} are accurately represented.}}
    \label{tab:example topics M10}
\end{table}
When randomly drawing training samples from the corpus, it might happen that we fail to draw one document from each class. This leads to the model detecting as any topics as we have classes in our training data. However, only labeling 40 documents already leads to 8.8 detected topics on average over 5 runs.

\begin{figure}[H]
    \centering
    \includegraphics[scale=0.3]{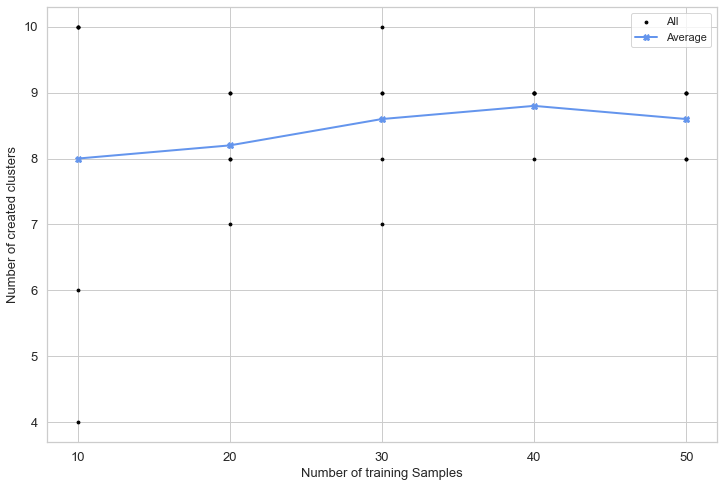}
    \caption{\small{Average number of extracted topics per labeled training samples. This is only for the completely randomly drawn labels.}}
    \label{fig:n_topics_M10}
\end{figure}

For the \textit{M10} dataset we have a slightly skewed document-topic distribution. This is also captured by the model, although the underrepresented classes are slightly more often predicted than they are truly prevalent. This could be due to the fact, that the training data does not accurately depict the true document-topic distribution prevalent in the dataset but is equally distributed over all classes\footnoteref{foot:note1}.

\begin{figure}[H]
    \centering
    \includegraphics[scale=0.5]{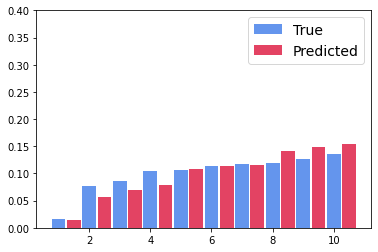}
    \caption{\small{Predicted topic distribution vs. true topic distribution. Results are achieved with DCTE on two labeled samples per class.}}
    \label{fig:n_topicdist_M10}
\end{figure}

\end{document}